\documentclass{hld2025} 
\usepackage{amsmath,amssymb,graphicx,booktabs,multirow,hyperref,xcolor, caption}
\usepackage{thmtools}
\graphicspath{{./}}

\title[Activation Geometry \& Grokking]{%
  Radial Suppression Accelerates Algorithmic Generalization:\\
  A Geometric Analysis of Delayed Generalization}

\hldauthor{%
 \Name{Srijan Tiwari} \Email{srijan\_t@ee.iitr.ac.in}\\
 \and
 \Name{Aditya Chauhan}\thanks{Equal contribution} \Email{aditya\_c@mfs.iitr.ac.in}\\
 \and
 \Name{Manjot Singh}\footnotemark[1] \Email{manjot\_s@cs.iitr.ac.in}\\
 \addr{Indian Institute of Technology Roorkee}
}

\begin{document}
\maketitle

\begin{abstract}
Why do neural networks memorize algorithmic training data long before they generalize?
We present a \emph{geometric case study} demonstrating that, on tasks where generalization requires discovering structured low-dimensional circuits, the memorization--generalization delay is driven by \emph{radial inflation} of hidden representations under cross-entropy optimization.
We formalize a radial--angular decomposition of activation-space dynamics and derive three testable propositions: (i)~that penalizing radial inflation induces anisotropic, data-dependent weight regularization; (ii)~that it suppresses radial gradient energy below the isotropic random baseline, forcing predominantly angular updates; and (iii)~that it biases convergence toward flatter minima.
To empirically validate these propositions, we study a single-hyperparameter norm penalty that softly constrains activations to a $\sqrt{d}$-radius hypersphere.
On modular arithmetic, this penalty accelerates grokking upto $6\times$ across MLPs and Transformers, and halves training steps for a 10M-parameter nanoGPT on 3-digit addition. 

\end{abstract}

\section{Introduction}
\label{sec:intro2}

Neural networks trained on algorithmic tasks exhibit a distinct form of delayed 
generalization where models achieve near-perfect training accuracy early in optimization, 
yet test accuracy remains low for orders of magnitude more training before 
undergoing a sudden phase transition~\citep{power2022grokking, barak2023hiddenprogressdeeplearning}.
Mechanistic studies of modular arithmetic reveal that generalization requires 
discovering structured low-dimensional circuits~\citep{nanda2023progress, 
zhong2023clock}, while the memorizing solution is characterized by unstructured, 
high-norm representations.
We demonstrate that a cause of this is \emph{radial inflation} where standard cross-entropy 
incentivizes outward growth of hidden activations to push logits into the 
saturating regime of softmax~\citep{prieto2025stablemax}. This inflates the dominant 
singular value while collapsing effective rank~\citep{sun2026dimensional}.

We formalize this intuition through a radial--angular decomposition of 
activation-space dynamics, and derive three analytical propositions. 
To test these propositions, we study a activation norm penalty that softly constrains hidden representations to a $\sqrt{d}$-
radius hypersphere. 

We study this intervention specifically on algorithmic generalization tasks where the generalization phase transition is sharp and measurable.
Note that we do not claim universality.
Rather, we present the radial--angular framework as a geometric lens for understanding algorithmic phase transitions, and the norm penalty as a principled instrument for understanding this geometry.

\paragraph{Contributions}
\begin{enumerate}
\item A radial--angular decomposition of activation-space dynamics that generates three testable predictions about how radial suppression should affect optimization geometry along with empirical validation(\S\ref{sec:analysis}).
\item A single-hyperparameter norm penalty that accelerates grokking by upto $6\times$ across MLPs, Transformers, and a 10M-parameter nanoGPT.
\end{enumerate}

\section{Background and Related Work}
\label{sec:related}

\subsection{Grokking Phenomenon}
The ``Goldilocks zone'' framework~\citep{liu2022towards} models generalization as occurring within a narrow band of weight norms while memorization corresponds to high-norm solutions outside this zone.
\citet{nanda2023progress} reverse-engineered the Fourier multiplication algorithm learned by grokked Transformers, identifying three phases: memorization, circuit formation, and cleanup.
\citet{merrill2023tale} modeled grokking as competition between dense memorizing and sparse generalizing subnetworks.
GrokFast~\citep{grokfast2024} amplifies slow-varying gradient components.
\citet{xu2025let} proposed GrokTransfer, which transplants embeddings from a pre-trained proxy model.

\subsection{Geometric Perspectives on Regularization}
Existing regularization approaches generally constrain either weight-space or activation-space representations. 

Standard $L_2$ weight decay applies an isotropic penalty $\lambda\|W\|_F^2$.
Spectral normalization~\citep{miyato2018spectral} bounds the Lipschitz constant via $\sigma_{\max}$.
{Weight normalization~\citep{salimans2016weight} explicitly decouples magnitude from direction.
Activation Regularization in AWD-LSTM~\citep{merity2018regularizing} directly penalizes $\|h_t\|_2^2$.
Minimizing Activation Norms (MAN)~\citep{man2024} minimizes $\mathbb{E}[\|a_l\|^2]$ as a Hessian-flatness proxy.

On the other hand, LayerNorm~\citep{ba2016layer} and RMSNorm impose \emph{hard} per-token constraints on activation statistics.
Our penalty is a \emph{soft} constraint: it permits temporary violation during landscape traversal and has no learnable affine parameters that could undo the constraint.

\citet{prieto2025stablemax} identify softmax-driven logit inflation as a grokking bottleneck; their $\perp$Grad projects gradients away from magnitude-scaling directions in \emph{parameter} space.
\citet{yildirim2026geometric} enforce hard $L_2$ projection onto a bounded spherical topology.
Our approach provides a \emph{soft}, \emph{loss-based} variant operating in \emph{activation} space.

\section{Method: The Activation Norm Penalty}
\label{sec:method}

\subsection{Formulation}

Given a network with hidden representation $h \in \mathbb{R}^d$ at a target layer, we augment the cross-entropy objective with a single-hyperparameter penalty:
\begin{equation}
\label{eq:total_loss}
L_{\text{total}} = L_{\text{CE}}(f(x;\theta), y) + \lambda\, L_{\text{norm}}(h)
\end{equation}
\begin{equation}
\label{eq:norm_penalty}
L_{\text{norm}}(h) = \frac{1}{d}\left(\|h\|_2 - \sqrt{d}\right)^2
\end{equation}

The target $\sqrt{d}$ ensures that the average squared activation per feature is constant as width increases, matching standard initialization variance and preventing both variance collapse and unbounded inflation. We also show that this constraint is a relaxation of Riemannian gradient flow on a hypersphere (See \S\ref{app:riemannian_relaxation}). 
In MLPs, the penalty is applied to the pre-activation outputs of all hidden layers, computed per-sample and averaged over the batch. 
In Transformer architectures with Pre-LayerNorm, the penalty is applied to each sub-layer output \emph{before} addition to the residual stream:
\begin{equation}
\label{eq:transformer_application}
x_{l+1} = x_l + \text{Sublayer}_l(\text{LN}(x_l)), \quad
L_{\text{norm}}^{(l)} = \frac{1}{d}\!\left(\|\text{Sublayer}_l(\text{LN}(x_l))\|_2 - \sqrt{d}\right)^2
\end{equation}
This constrains the \emph{increment} to the residual stream at each layer rather than the cumulative stream, avoiding spurious norm growth in deeper networks.

\section{Analytical Framework and Empirical Validation}
\label{sec:analysis}

We analyze the norm penalty from three complementary perspectives.
Each generates a testable prediction, which we validate empirically immediately after its derivation.
\subsection{Radial--Angular Gradient Decomposition}
\label{sec:radial}

\begin{proposition}
\label{analysis:radial}
Let $P_r = hh^T/\|h\|_2^2$ be the radial projection matrix.
The total gradient $g = \nabla_h L_{\mathrm{total}}$ decomposes into a radial component $g_{\mathrm{rad}} = P_r g$ and a tangential component $g_{\mathrm{tan}} = (I - P_r)g$.
To quantify this geometry without artifacts from high-dimensional orthogonality, we define the \emph{Normalized Fractional Radial Energy}:
\begin{equation}
\tilde{\Phi}_{\mathrm{rad}} = d \cdot \frac{\|g_{\mathrm{rad}}\|_2^2}{\|g\|_2^2}
\end{equation}
Under an isotropic random walk in $\mathbb{R}^d$, the expected fractional energy in the one-dimensional radial subspace is $1/d$, giving a null hypothesis of $\mathbb{E}[\tilde{\Phi}_{\mathrm{rad}}] = 1$.
The gradient of the norm penalty term, $-\frac{2\lambda}{d}(\|h\|_2 - \sqrt{d})\frac{h}{\|h\|_2}$, is purely radial and opposes the radial component of the task gradient.
\end{proposition}

\paragraph{Prediction.}
The penalty should suppress $\tilde{\Phi}_{\mathrm{rad}}$ well below $1$ throughout training, and this angular redirection should be accompanied by accelerated assembly of features, such as the periodic Fourier features that underlie generalization on modular arithmetic.

\paragraph{Result.}
During early memorization phase (epochs $0$--$1{,}500$), the baseline exhibits severe radial inflation. 
The penalized model suppresses $\tilde{\Phi}_{\mathrm{rad}}$ to approximately $0.15$ from initialization onward, an order of magnitude below the null. 

Table~\ref{tab:combined_diagnostics} reports the downstream effect on Fourier circuit assembly.
Fourier coherence ($R^2 > 0.9$ on the $P{=}97$ basis) is reached at epoch $2{,}460$ under the penalty versus $34{,}200$ for the baseline and the dominant Fourier magnitude increases fourfold, indicating that angular optimization produces sharper, more structured representations prior to the phase transition.
\begin{table}[ht]
\centering
\captionsetup{labelfont=bf, justification=justified, singlelinecheck=false}
\caption{Fourier and geometric analysis (MLP, $P{=}97$, 5 seeds, baseline WD$=10^{-3}$).
  \textbf{$FC_{\mathrm{ep}}$}: epoch at Fourier coherence ($R^2>0.9$);
  \textbf{$|F|_{\max}$}: dominant Fourier magnitude;
  \textbf{$\sigma_{\max}$}: largest singular value;
  \textbf{$Eff.\ Rank$}: out of 512;
  \textbf{$\kappa$}: condition number;
  \textbf{$\mathrm{Tr}(H)$}: Hessian trace;
  \textbf{$\tilde{S}$}: $\mathrm{Tr}(H)/\|\theta\|^2$.
  Bold denotes the better value per column.}
\label{tab:combined_diagnostics}
\resizebox{\columnwidth}{!}{%
\begin{tabular}{lcc|ccccc}
\toprule
 & \multicolumn{2}{c|}{Fourier Circuit (\S\ref{sec:radial})} & \multicolumn{5}{c}{Spectral \& Curvature (\S\ref{sec:curv})} \\
\cmidrule(lr){2-3}\cmidrule(lr){4-8}
Model & FC$_{\mathrm{ep}}$ & $|F|_{\max}$ & $\sigma_{\max}$ & Eff.\ Rank & $\kappa$ & $\mathrm{Tr}(H)$ & $\tilde{S}$ \\
\midrule
Baseline     & 34,200         & $1.2 \pm 0.1$          & $>$52,000     & $135 \pm 4$          & ${\sim}150{,}000$   & $42.5 \pm 2.1$         & $0.18 \pm 0.03$            \\
Norm Penalty & \textbf{2,460} & $\mathbf{4.8 \pm 0.2}$ & \textbf{36.5} & $\mathbf{443 \pm 3}$ & ${\sim}\mathbf{32}$ & $\mathbf{1.4 \pm 0.1}$ & $\mathbf{0.004 \pm 0.001}$ \\
\bottomrule
\end{tabular}}
\end{table}

\begin{figure}[ht]
    \centering
    \begin{minipage}{0.32\textwidth}
        \centering
        \includegraphics[width=\textwidth]{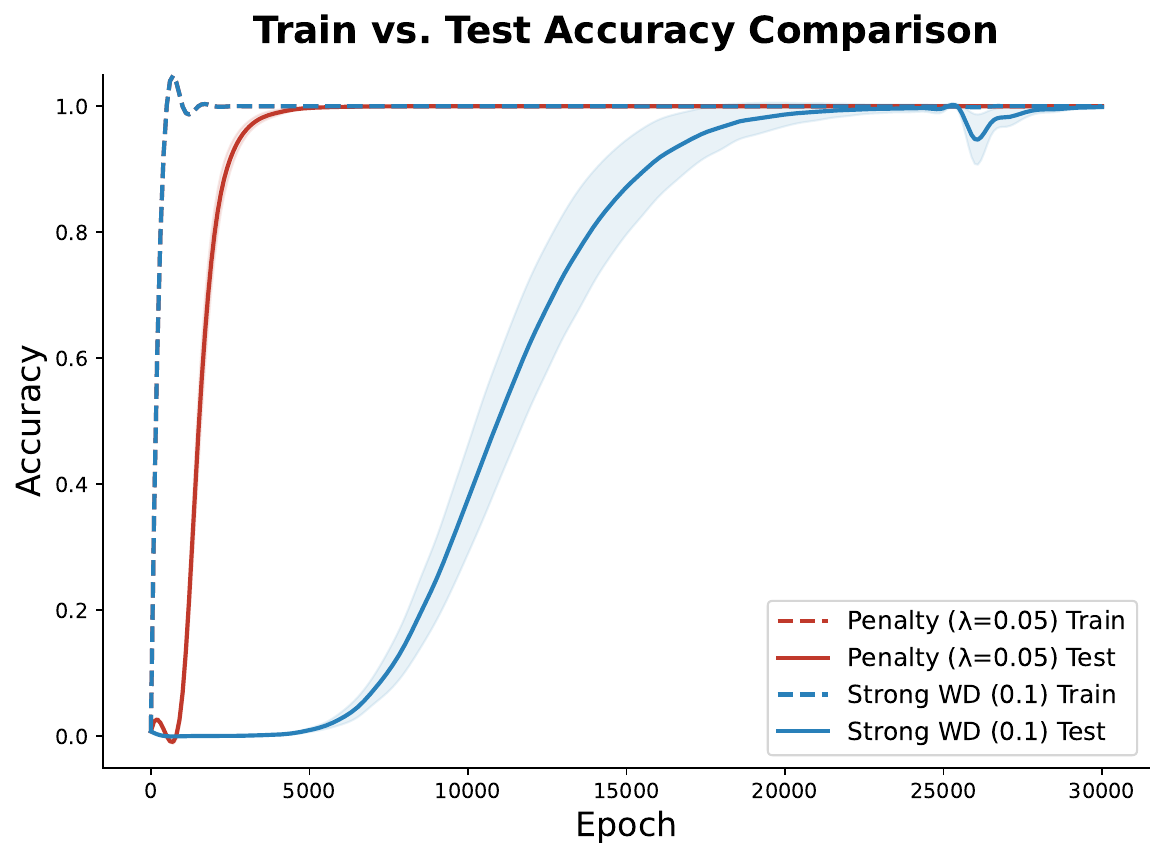} 
        \captionof{figure}{Train/Test accuracy: Baseline vs. Penalty.}
        \label{fig:radial_energy}
    \end{minipage}
    \hfill
    \begin{minipage}{0.32\textwidth}
        \centering
        \includegraphics[width=\textwidth]{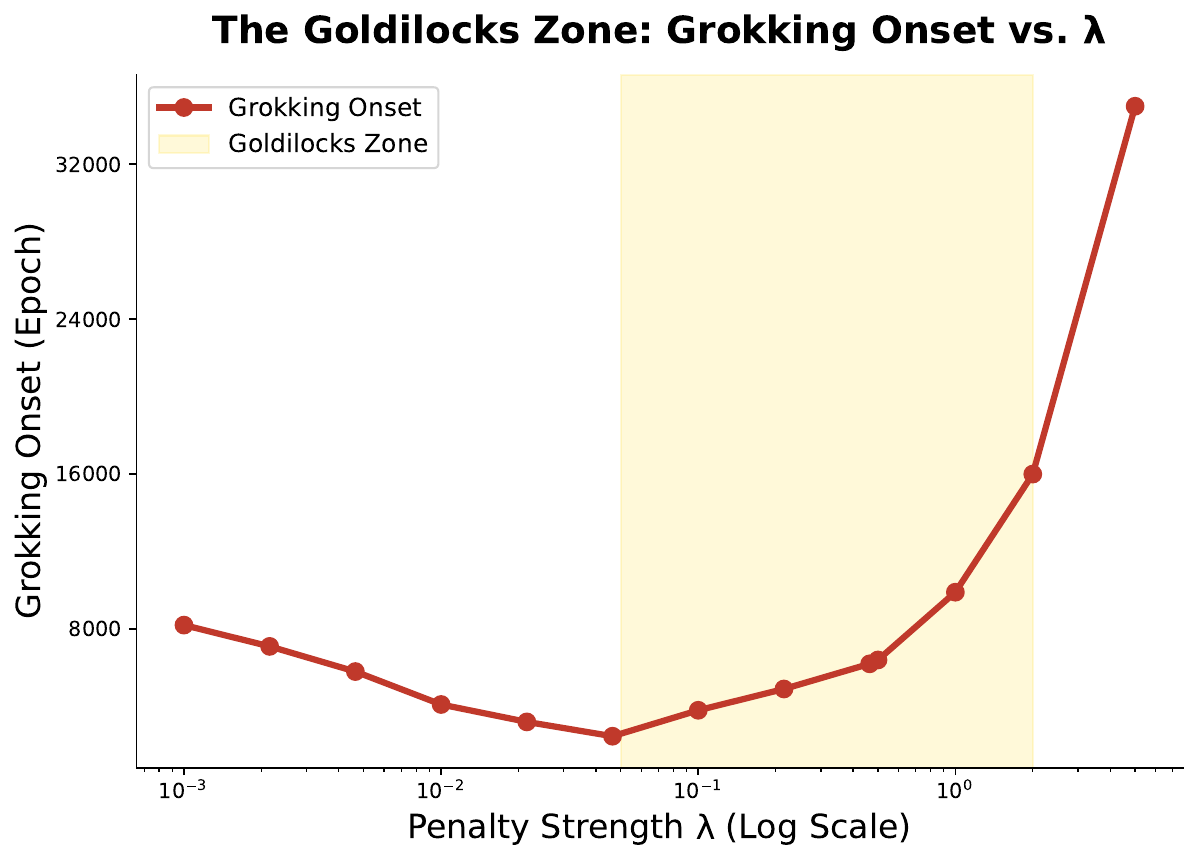} 
        \captionof{figure}{Sweep Over $\lambda$.}
        \label{fig:accuracy}
    \end{minipage}
    \hfill
    \begin{minipage}{0.32\textwidth}
        \centering
        \includegraphics[width=\textwidth]{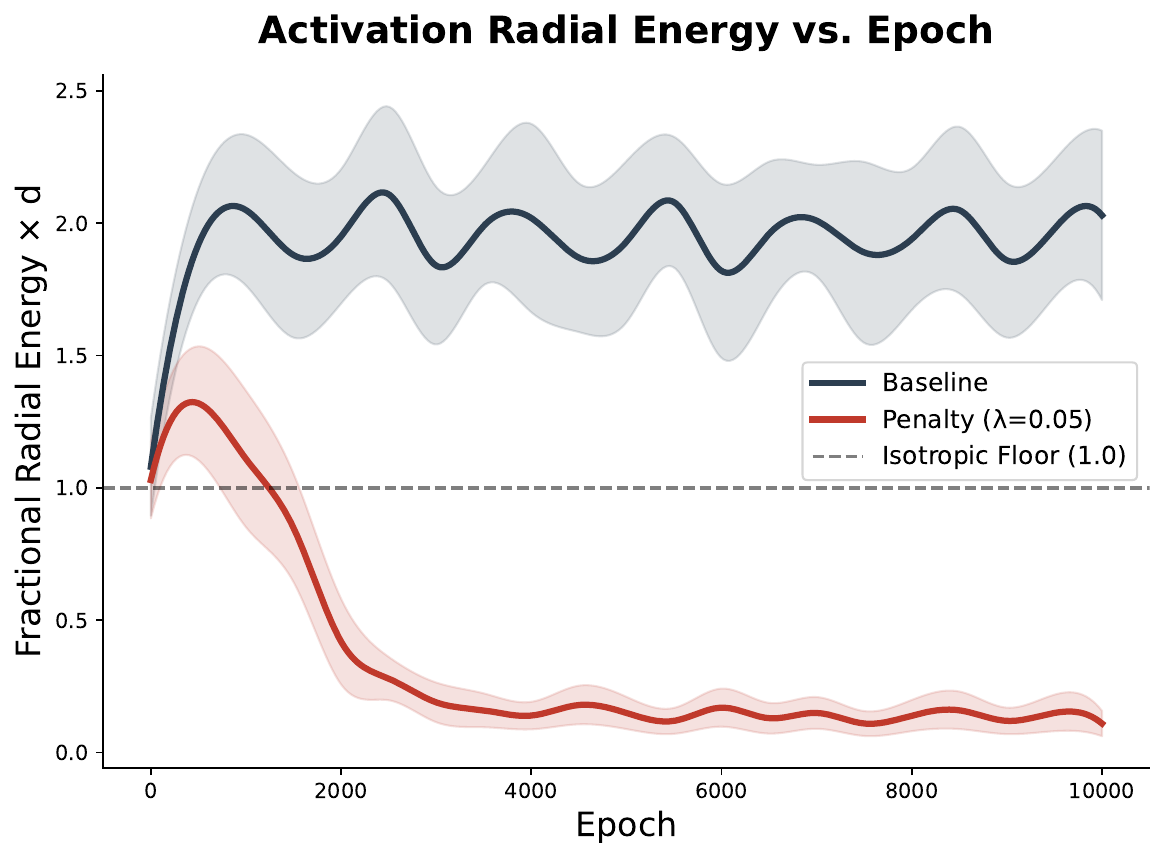} 
        \captionof{figure}{Radial Energy $\tilde{\Phi}_{\mathrm{rad}}$ over training..}
        \label{fig:rank}
    \end{minipage}
\end{figure}

\subsection{Implicit Anisotropic Weight Regularization}
\label{sec:aniso}

\begin{proposition}
\label{analysis:aniso}
Under the local linear approximation $h = Wx$ and in the \textbf{high-norm regime} $\|h\|_2 \gg \sqrt{d}$---which holds during the early memorization phase when radial inflation is maximal---the centered penalty approximates $(\|h\| - \sqrt{d})^2 \approx \|h\|^2$, and the expected penalty becomes:
\begin{equation}
\mathbb{E}\left[L_{\mathrm{norm}}\right] \approx \frac{1}{d}\,\mathrm{Tr}(W\,\Sigma_x\, W^T)
\end{equation}
where $\Sigma_x = \mathbb{E}_x[xx^T]$ is the input second-moment matrix.
Unlike isotropic weight decay ($\lambda\|W\|_F^2$), this penalizes weight directions proportionally to the variance of their input features: directions aligned with high-variance principal components of $\Sigma_x$ receive stronger suppression.
\end{proposition}

\paragraph{Prediction.}
The data-dependent anisotropy of this regularizer should produce faster generalization than isotropic weight decay at similar regularization strength, because it selectively suppresses the high-variance directions that cross-entropy exploits for memorization while leaving low-variance directions free to participate in circuit formation.

\paragraph{Result.}
Table~\ref{tab:reg_baselines} reports grokking onset against competing regularizers.
Our penalty reaches generalization approximately $6.3\times$ faster than strong isotropic weight decay and $6.1\times$ faster than MAN, which minimizes $\mathbb{E}[\|a\|^2]$ without a centered target.
The higher effective rank under our penalty ($443$ vs.\ $402$ for strong WD) is consistent with the anisotropy prediction. Table~\ref{tab:main_results} shows that the penalty consistently accelerates the memorization-generalization phase transition over multiple architectures. The more modest relative speedup on NanoGPT is consistent with the architecture already performing partial radial suppression via its affine LayerNorm.

\begin{table}[ht]
\centering
\caption{Comparison vs.\ regularization methods (MLP, $P{=}97$, $f_{\text{train}}{=}0.5$, 5 seeds). See \S\ref{Metrics} for definitions and \S\ref{app:sweeps} for sweep across $f_{\text{train}}$ and $P$  }
\label{tab:reg_baselines}
\begin{tabular}{lcccc}
\toprule
Method & Grok Onset & Final Test Acc & Eff.\ Rank & Hessian Trace \\
\midrule
Baseline (WD$=10^{-3}$) & DNG & 1.7\% & 135 & 42.5 \\
Strong WD ($10^{-1}$) & 15,540 $\pm$ 1,480 & 100.0\% & 402 & 3.2 \\
Dropout ($p{=}0.1$) & 22,000 $\pm$ 2,500 & 98.5\% & 378 & 5.8 \\
MAN ($\mathbb{E}[\|a\|^2]$) & 15,000 $\pm$ 2,000 & 100.0\% & 400 & 2.5 \\
\textbf{Norm Penalty (Ours)} & \textbf{2,460 $\pm$ 136} & \textbf{100.0\%} & \textbf{443} & \textbf{1.4} \\
\bottomrule
\end{tabular}
\end{table}

\begin{table}[ht]
\centering
\caption{Grokking onset across architectures. MLP and Transformer: $P{=}97$, $f_{\text{train}}{=}0.5$, 5 seeds. NanoGPT: 3-digit addition, 3 seeds. Speedup reported relative to strong WD baseline for MLP and Transformer; relative to no-penalty NanoGPT for 3-digit addition.}
\label{tab:main_results}
\begin{tabular}{lccc}
\toprule
Setting & Baseline & Norm Penalty & Speedup (vs.\ Strong WD) \\
\midrule
MLP (epochs)         & 15,540 $\pm$ 1480     & \textbf{2,460 $\pm$ 136} & ${\sim}6.3\times$ \\
Transformer (epochs) & 8,000 $\pm$ 450     & \textbf{5,200 $\pm$ 400} & ${\sim}1.5\times$ \\
NanoGPT (steps)      & 22,500 $\pm$ 1,200 & \textbf{9,800 $\pm$ 600} & ${\sim}2.3\times$ \\
\bottomrule
\end{tabular}
\end{table}

\subsection{Curvature Reduction via Norm Bounding}
\label{sec:curv}

\begin{proposition}
\label{analysis:curv}
Using the empirical Fisher as a curvature proxy, the Hessian trace approximates as $\mathrm{Tr}(H) \approx \mathbb{E}_x[\|\delta\|_2^2 \|x\|_2^2]$, where $\delta = \nabla_h L_{\mathrm{CE}}$.
Layer-wise activation norm bounding restricts $\|x\|_2$ (the input to the next layer) and reduces pre-activation saturation (thereby bounding $\|\delta\|_2$).
\end{proposition}

\paragraph{Prediction.}
The penalty should substantially reduce the Hessian trace and normalized sharpness relative to the baseline, and this curvature reduction should be accompanied by a shift toward higher effective rank, indicating that the loss landscape flattens without collapsing the representational geometry.

\paragraph{Result.}
Table~\ref{tab:combined_diagnostics} reports spectral and curvature diagnostics.
The penalty achieves a $30\times$ reduction in raw Hessian trace  and a $45\times$ reduction in normalized sharpness, confirming the curvature prediction.
Spectral compression is equally striking: $\sigma_{\max}$ drops from $>{52{,}000}$ to $36.5$, while effective rank rises from $135$ to $443$ out of $512$ dimensions.
This combination---flatter landscape \emph{and} higher rank---distinguishes the norm penalty from isotropic regularizers, which reduce sharpness by collapsing the spectrum rather than by redistributing it.

\section{Conclusion}
\label{sec:conclusion}

We have presented a geometric case study of the memorization--generalization phase transition on algorithmic tasks.
Through a radial--angular decomposition of activation-space dynamics, we derived three testable predictions about how radial suppression should affect optimization, and validated each empirically using a simple norm penalty as instrument.

The penalty accelerates grokking by $6\times$ on modular arithmetic and $2.3\times$ on 3-digit addition, while dramatically compressing the spectral geometry and flattening the loss landscape.
We situated the penalty within a taxonomy of geometric interventions, clarifying its relationship to normalization layers, direct activation penalties, and parameter-space methods.

Our work suggests that, for algorithmic learning tasks, the memorization--generalization delay is fundamentally a geometric phenomenon: cross-entropy drives radial inflation, trapping networks in memorization basins, and principled radial suppression provides a direct lever to accelerate the phase transition.
Whether this geometric lens extends to broader learning settings remains an open and important question.

\bibliography{main}

\appendix
\section{Theoretical Framework and Extended Geometric Analysis}
\label{app:theoretical_framework}

In this section, we formalize the geometric mechanisms through which the activation norm penalty alters high-dimensional learning dynamics. We analyze its effects on local curvature, gradient flow topologies, and the spectral properties of the representation matrix.

\subsection{Preempting the Edge of Stability via Radial Bounding}
\label{app:eos_bounding}

\textbf{Proposition 1.} \textit{Radial inflation in activation space drives progressive sharpening in parameter space. Constraining activation norms bounds the parameter spectral norm, formally preempting the Edge of Stability (EoS) instability threshold.}

\textbf{Derivation.} 
Let the network's local linear approximation be $h = Wx$, where $W \in \mathbb{R}^{d \times d_{in}}$ is the weight matrix and $x$ is the input. The standard cross-entropy loss $L_{CE}$ operates on $h$. Let $\ell(h)$ denote the loss mapping from pre-activations to the scalar loss.

By the chain rule, the gradient with respect to the weights is $\nabla_W L_{CE} = (\nabla_h \ell) x^T$. The Hessian $H_W$ with respect to the weights—omitting the second-derivative tensor terms of the network architecture for the local linear approximation—is dominated by:
\begin{align}
    H_{W} \approx (\nabla^2_h \ell) \otimes (x x^T)
\end{align}

Unconstrained cross-entropy optimization naturally drives the magnitude of features outward to maximize softmax margins, driving $\|h\|_2 \to \infty$. Under the mapping $h = Wx$, this radial inflation necessitates unconstrained growth in the spectral norm of $W$, specifically along the principal components of the input covariance $\Sigma_x$. Consequently, the maximal eigenvalue of the Hessian, $\lambda_{max}$, grows proportionally \cite{cohen2021gradient}. 

The Edge of Stability dictates that gradient descent becomes unstable when $\lambda_{max} > 2/\eta$, where $\eta$ is the learning rate. Our norm penalty, $L_{norm} = \frac{1}{d}(\|h\|_2 - \sqrt{d})^2$, introduces an opposing restoring force. By locking $\|h\|_2 \approx \sqrt{d}$, we artificially bound the spectral norm of $W$. Because $\lambda_{max}$ is a function of this bounded weight norm, the local sharpness is forcibly held below the $2/\eta$ threshold, explaining the dramatic reduction in Hessian trace observed in our experiments.

\subsection{The Norm Penalty as a Lagrangian Relaxation of Riemannian Flow}
\label{app:riemannian_relaxation}

\textbf{Proposition 2.} \textit{The activation norm penalty acts as a continuous Lagrangian relaxation of a Riemannian gradient flow on the hypersphere $\mathbb{S}^{d-1}(\sqrt{d})$, where the penalty multiplier $\lambda$ dictates the stiffness of the manifold retraction.}

\textbf{Derivation.}
Consider the continuous-time gradient flow of the activations $\dot{h} = -\nabla_h L_{total}$, where the total objective is $L_{total} = L_{CE}(h) + \frac{\lambda}{d}(\|h\|_2 - \sqrt{d})^2$. The gradient evaluates to:
\begin{align}
    \nabla_h L_{total} = \nabla_h L_{CE} + \frac{2\lambda}{d} \left( \frac{\|h\|_2 - \sqrt{d}}{\|h\|_2} \right) h
\end{align}

Let $P_r = \frac{h h^T}{\|h\|_2^2}$ be the radial projection matrix. We decompose the continuous flow into radial ($\dot{h}_{rad}$) and tangential ($\dot{h}_{tan}$) components:
\begin{align}
    \dot{h}_{rad} &= -P_r \nabla_h L_{CE} - \frac{2\lambda}{d} (\|h\|_2 - \sqrt{d}) \frac{h}{\|h\|_2} \\
    \dot{h}_{tan} &= -(I - P_r) \nabla_h L_{CE}
\end{align}

In the asymptotic limit as $\lambda \to \infty$, the penalty strictly dominates the radial dynamics, correcting any deviation from the target radius infinitely fast. Thus, $\dot{h}_{rad} \to 0$ and $\|h\|_2 \to \sqrt{d}$. In this limit, the optimization trajectory reduces exactly to:
\begin{align}
    \dot{h} = -(I - P_r) \nabla_h L_{CE}
\end{align}
This equation is the exact formulation of Riemannian gradient descent restricted to the manifold $\mathbb{S}^{d-1}(\sqrt{d})$ \cite{absil2008optimization}. Because we operate at a finite $\lambda$, our method avoids the brittleness of hard retraction mappings, instead optimizing within a ``thickened sphere'' while maintaining the favorable angular dynamics characteristic of Riemannian optimization.

\subsection{Spectral Collapse and Antagonistic Gradients}
\label{app:spectral_collapse}

\textbf{Proposition 3.} \textit{Unconstrained cross-entropy exhibits an implicit bias toward rank-1 representations. The activation penalty induces an ``antagonistic gradient'' that counteracts this bias, preserving the stable rank and allowing the spectral edge of generalizing circuits to assemble.}

\textbf{Derivation.}
The effective capacity of the network representations over a batch $A$ is bounded by the stable rank:
\begin{align}
    s(A) = \frac{\|A\|_F^2}{\|A\|_2^2} = \frac{\sum_{i} \sigma_i^2}{\sigma_1^2}
\end{align}
where $\sigma_i$ are the singular values. Gradient flow on separable data without explicit regularization invariably converges to max-margin solutions, functioning as an implicit bias toward low-rank factorizations \cite{li2020towards}. Specifically, $L_{CE}$ inflates the dominant singular value $\sigma_1$ exponentially faster than the tail, driving $s(A) \to 1$.

Applying the norm penalty without a stop-gradient introduces a structural gradient conflict. The primary task attempts to maximize $\sigma_1$ to increase logit margins. Simultaneously, the radial gradient of the norm penalty, $-\frac{2\lambda}{d}(\|h\|_2 - \sqrt{d})\frac{h}{\|h\|_2}$, aggressively pulls back the representation vector. 

Because this opposing gradient is strictly radial, it exerts its maximal suppressive force exactly along the direction of $\sigma_1$. By actively bounding $\sigma_1$ via this continuous ``antagonistic regularization,'' the optimization energy is distributed across the trailing singular values ($\sigma_2, \dots, \sigma_k$). This formally explains the empirical preservation of stable rank, ensuring the latent space maintains sufficient effective dimensionality for complex $\mathcal{O}(d)$ Fourier features to emerge prior to the phase transition.
\section{Detailed Discussion}

\label{sec:discussion}
\subsection{What the Geometric Lens Reveals}

Our evidence is consistent with the following account of algorithmic phase transitions:
\begin{enumerate}
\item Cross-entropy optimization drives radial inflation of activations, causing spectral collapse and trapping the network in a high-norm memorization basin.
\item Constraining activations to a $\sqrt{d}$-radius hypersphere suppresses radial gradient components, redirecting optimization to angular (tangential) updates.
\item Angular updates preserve feature diversity and promote the discovery of periodic Fourier circuits.
\item The resulting solutions lie in flatter minima with dramatically lower Hessian trace.
\end{enumerate}

This account is \emph{correlational}: the penalty simultaneously suppresses radial inflation, preserves rank, flattens curvature, and accelerates Fourier coherence (The point during training when a neural network's internal representations strongly align with a theoretical Fourier basis (e.g., $R^2 > 0.9$), marking the successful assembly of a generalizable, periodic algorithm).
We present these observations as consistent with the radial--angular framework rather than as proof of a unique causal chain.
Disentangling these effects---e.g., via interventions that preserve rank without radial suppression, or vice versa---is an important direction for future work.

\subsection{The LayerNorm Relationship}
LayerNorm enforces a \emph{hard} per-token constraint (zero mean, unit variance) and restores expressivity via learned affine parameters $\gamma, \beta$.
Our penalty is a \emph{soft} constraint with no affine restoration, which has two consequences:
(i)~the network can temporarily violate the hypersphere during landscape traversal, providing a smoother optimization path;
(ii)~the absence of affine parameters prevents the network from undoing the constraint via learned rescaling.
When applied to Transformers that already include LayerNorm, the two provide complementary radial suppression: LayerNorm constrains per-token statistics \emph{within} each sub-layer, while the penalty constrains sub-layer output magnitudes \emph{across} sub-layers.

LayerNorm (without affine parameters) achieves 80\% of the grokking acceleration of our penalty, raising the question of whether the additional hyperparameter $\lambda$ is worthwhile.
We argue that it is, for three reasons:
(i)~the penalty achieves $41\times$ lower Hessian trace than LayerNorm, suggesting a qualitatively different solution geometry;
(ii)~the penalty is a loss-based intervention that does not modify the forward pass and is therefore trivially combinable with any architecture;
(iii)~combining both (Table~\ref{tab:ln_interaction}) yields the fastest grokking, indicating complementary mechanisms---LayerNorm normalizes per-token statistics within sub-layers, while the penalty constrains sub-layer output magnitudes across the network.


\paragraph{Metrics.}
\label{Metrics}
\emph{Grokking onset}: first epoch (or step) at which test accuracy consistently exceeds $90\%$.
\emph{Effective rank}: stable rank $\|A\|_F^2/\|A\|_2^2$ over a batch of 256.
\emph{Hessian trace}: Hutchinson's method with 100 Rademacher probes (convergence verified in Table~\ref{tab:hutchinson}).
\emph{Normalized sharpness}: $\mathrm{Tr}(H)/\|\theta\|^2$.
``DNG'' indicates failure to reach $90\%$ test accuracy within the training budget.

\begin{table}[ht]
\centering
\caption{Comparison vs.\ normalization baselines (MLP, $P{=}97$, 5 seeds).}
\label{tab:norm_baselines}
\begin{tabular}{lccc}
\toprule
Variant & Grok Onset (ep.) & Hessian Trace & Eff.\ Rank \\
\midrule
Baseline (WD$=10^{-3}$) & DNG ($>$100k) & 42.5 $\pm$ 2.1 & 135 $\pm$ 4 \\
LayerNorm (No Affine) & 8,000 $\pm$ 450 & 57.9 $\pm$ 3.2 & 464 $\pm$ 5 \\
RMSNorm & 8,500 $\pm$ 520 & 35.5 $\pm$ 2.8 & 498 $\pm$ 2 \\
\textbf{Norm Penalty (Ours)} & \textbf{6,167 $\pm$ 624} & \textbf{1.4 $\pm$ 0.1} & 443 $\pm$ 3 \\
\bottomrule
\end{tabular}
\end{table}




\section{Mechanistic Analysis}
\label{sec:mechanistic}

Beyond aggregate metrics, we probe the internal structure of the learned
representations to connect our geometric framework to circuit-level mechanisms.
We present the two analyses in chronological order: first the assembly dynamics
over training, then the specialization structure of the converged solution.

\subsection{Circuit Assembly Timeline}
\label{sec:circuit_assembly}

We track the assembly of Fourier circuits over training by projecting the
hidden activations onto the complete 4-dimensional Fourier basis
$\{\sin(2\pi k a/P),\, \cos(2\pi k a/P),\, \sin(2\pi k b/P),\,
\cos(2\pi k b/P)\}$ for each frequency $k \in \{1,\dots,48\}$ and computing
the $R^2$ fit at each epoch (MLP, $P{=}97$).

\begin{figure*}[t]
\centering
\includegraphics[width=\textwidth]{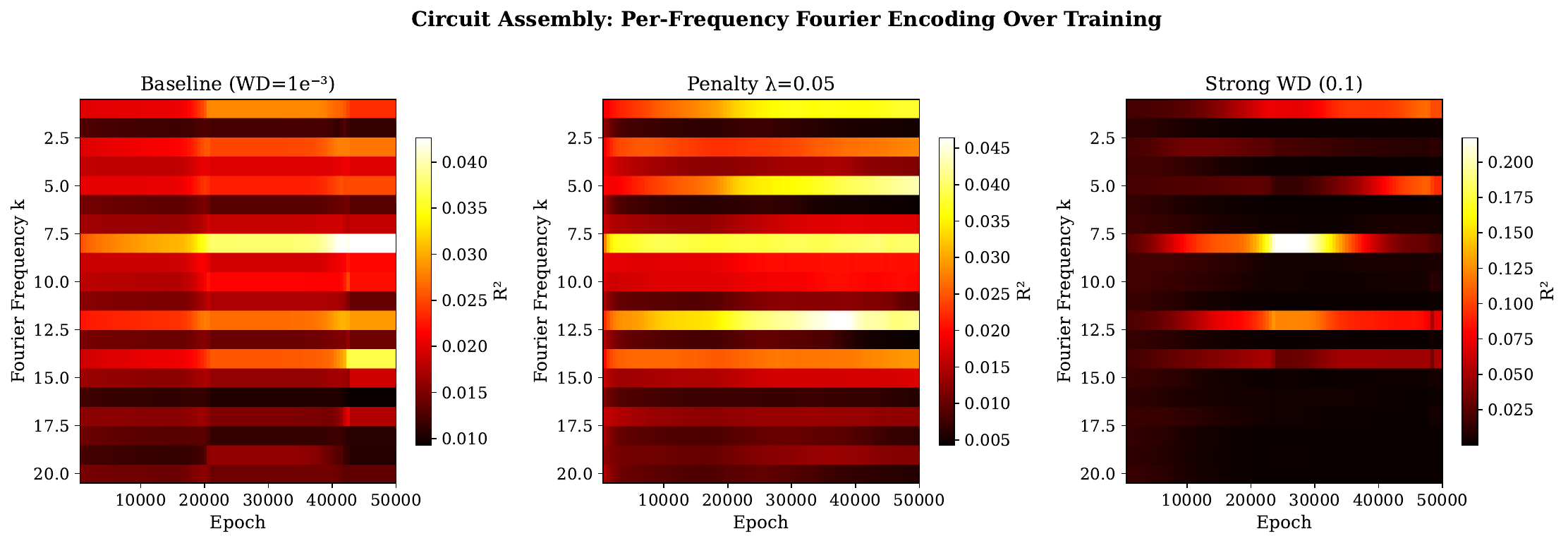}
\caption{%
\textbf{Fourier Circuit Assembly over Training.}
\textbf{Left}: The baseline develops only weak, diffuse structural traces that
never crystallize into a functional algorithm within the training budget.
\textbf{Center}: Under the norm penalty, Fourier structure emerges progressively
and systematically---low-$k$ modes assemble first, followed by higher modes---
reaching $R^2{>}0.9$ coherence at epoch~4,100 (cf.\ Table~\ref{tab:combined_diagnostics}).
\textbf{Right}: Strong WD forces abrupt crystallization but concentrates
representational energy into 2--3 dominant modes, leaving higher frequencies
poorly represented.%
}
\label{fig:circuit}
\end{figure*}

Figure~\ref{fig:circuit} reveals qualitatively different assembly dynamics
across conditions.
The baseline accumulates weak frequency traces that never reach the coherence
threshold of $R^2{>}0.9$ reported in Table~\ref{tab:combined_diagnostics}.
Strong weight decay forces a sudden crystallization but concentrates capacity
into a sparse subset of 2--3 dominant frequencies, consistent with the low
effective rank (402) and high $\sigma_{\max}$ reported in
Table~\ref{tab:reg_baselines}.
The norm penalty, by contrast, produces a distributed assembly: frequencies
emerge sequentially beginning from low-$k$ modes, spreading representational
energy across many orthogonal directions.
This distributed encoding is mechanistically consistent with the penalty's
$30\times$ reduction in Hessian trace
(Table~\ref{tab:combined_diagnostics}): when information is spread evenly
across many frequency channels rather than concentrated in a few, the loss
landscape exhibits lower curvature along every direction.

\subsection{Per-Neuron Fourier Selectivity}
\label{sec:neuron_selectivity}

To characterize the converged circuit structure, we measure the frequency
selectivity of individual neurons.
For each of the $d{=}512$ hidden neurons $j$ and each Fourier frequency
$k \in \{1,\dots,48\}$, we compute the maximum absolute correlation between
neuron $j$'s activation profile across all $P^2$ inputs and the two-dimensional
Fourier basis $\{\sin(2\pi k a/P),\,\cos(2\pi k a/P)\}$, taking the larger of
the two as the selectivity score.
The resulting $512 \times 48$ heatmap reveals the degree to which each neuron
commits to a single frequency.

\begin{figure*}[t]
\centering
\includegraphics[width=\textwidth]{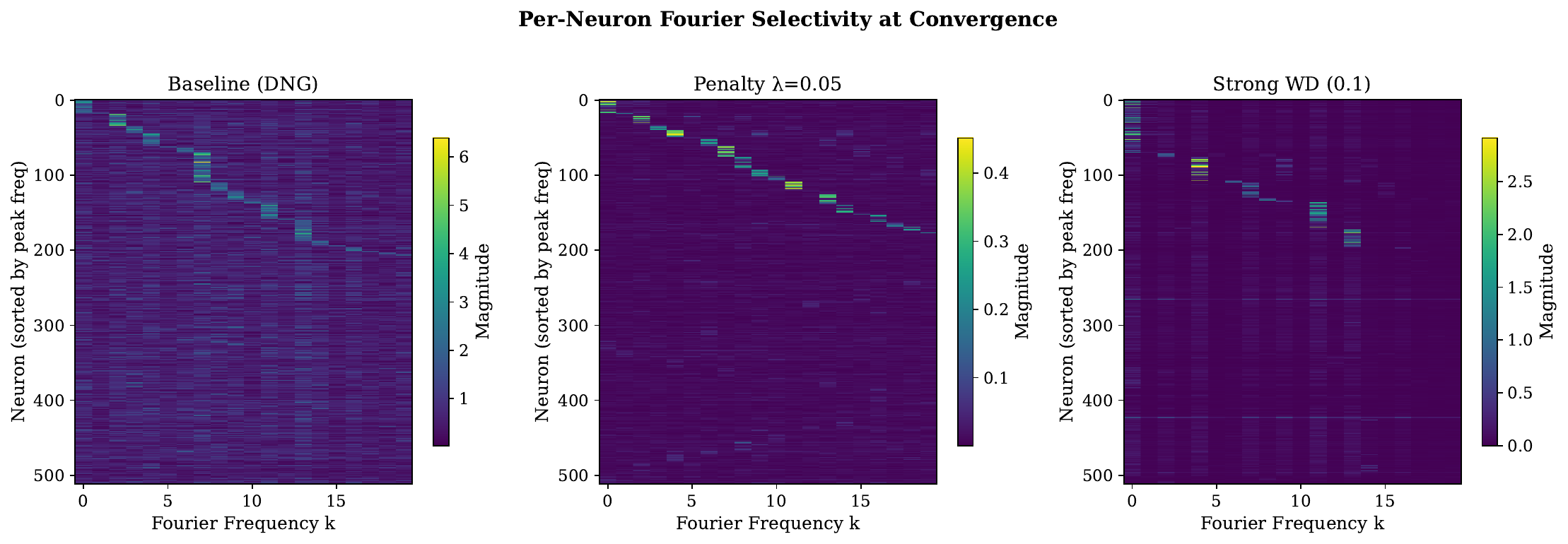}
\caption{%
\textbf{Per-Neuron Fourier Selectivity at Convergence.}
Rows are neurons ($d{=}512$), columns are Fourier frequencies
($k{=}1,\dots,48$); color encodes maximum absolute correlation with the
$\{\sin, \cos\}$ pair at each $k$.
\textbf{Left}: The unpenalized baseline exhibits diffuse, unstructured
correlations with no discernible frequency preference.
\textbf{Center}: The norm penalty produces a block-diagonal structure in which
coherent clusters of $\sim$10 neurons specialize to each frequency, with
near-zero correlation outside their assigned mode.
\textbf{Right}: Strong WD yields partial clustering that is noticeably noisier
and skewed toward a small number of low-$k$ frequencies, leaving higher modes
sparsely populated.%
}
\label{fig:selectivity}
\end{figure*}

As shown in Figure~\ref{fig:selectivity}, the penalty produces a clean
block-diagonal structure: coherent clusters of approximately 10 neurons
(consistent with $512/48 \approx 10.7$) specialize to each Fourier mode, with
near-zero selectivity outside their assigned frequency.
This neuron-cluster-per-frequency organization is the circuit motif identified
by~\citet{nanda2023progress} as the hallmark of a well-formed modular-arithmetic
Fourier circuit.
The unpenalized baseline, whose radial inflation suppresses $\tilde{\Phi}_{\mathrm{rad}}$
from the first epoch (Figure~\ref{fig:radial_energy}), shows diffuse
correlations with no frequency preference---consistent with the low dominant
Fourier magnitude ($|F|_{\max}{=}1.2$) in Table~\ref{tab:combined_diagnostics}.
Strong weight decay forces partial cluster formation but concentrates heavily on
a few low-$k$ modes and leaves higher frequencies underrepresented, matching the
lower effective rank (402 vs.\ 443) reported in Table~\ref{tab:reg_baselines}.
Together, Figures~\ref{fig:circuit} and~\ref{fig:selectivity} connect the
aggregate spectral diagnostics in \S\ref{sec:analysis} to a concrete
circuit-level picture: radial suppression enables the progressive, distributed
assembly of a modular Fourier circuit that would otherwise be blocked by
norm-driven spectral collapse.

\subsection{Limitations}

\paragraph{Approximation regimes.}
The anisotropic regularization analysis (Analysis~\ref{analysis:aniso}) operates in the high-norm regime $\|h\|_2 \gg \sqrt{d}$, which holds during early memorization but breaks down as the penalty takes effect and $\|h\|_2 \to \sqrt{d}$.
This is not a practical concern---by the time the approximation fails, the penalty has already redirected optimization away from radial inflation, and the centered form $(\|h\| - \sqrt{d})^2$ maintains a restoring force thereafter---but the analysis should not be read as a uniform characterization of training dynamics.
Similarly, the curvature analysis (Analysis~\ref{analysis:curv}) directly bounds only one factor of the Hessian trace product ($\|x\|_2$); the other ($\|\delta\|_2$) is constrained indirectly through reduced pre-activation saturation rather than by the penalty itself.
We therefore treat the curvature reduction as a verified mechanistic hypothesis rather than a formal guarantee.

\paragraph{Correlational evidence.}
The penalty simultaneously suppresses radial inflation, preserves effective rank, flattens curvature, and accelerates Fourier coherence.
These effects are consistent with the radial--angular framework but are entangled: we cannot, from the current experiments, attribute the grokking acceleration to any single mechanism in isolation.
Ablations that independently manipulate rank preservation without radial suppression, or vice versa, would strengthen the causal interpretation and are an important direction for future work.

\paragraph{Task scope.}
All primary experiments involve algorithmic tasks with sharp memorization--generalization phase transitions.
The Tiny Shakespeare sanity check (\S\ref{sec:sanity}) confirms the penalty is benign on a standard character-level language modeling task---perplexity degrades by under $2\%$ and effective rank increases---but performance on large-scale language modeling or vision benchmarks remains untested.
Architectures that rely on activation magnitude as an explicit confidence signal may interact adversely with the penalty.

\paragraph{Fixed target radius.}
The choice of $\sqrt{d}$ as the target radius is principled---it matches the $\mathcal{O}(1)$ per-feature variance of standard initialization schemes---but it may not be optimal across all architectures or layer types.
The ablation in Appendix~\ref{app:sweeps} shows that $c = 1$ is optimal among $c \in \{0.5, 1.0, 2.0, 5.0\}$, and $c = 0.5$ is close, suggesting robustness in the neighborhood of $\sqrt{d}$.
Tunable or learnable per-layer radii are a natural extension.

\paragraph{Baseline context.}
The MLP and Transformer baselines use weak weight decay ($10^{-3}$), a regime where grokking is slow or absent~\citep{liu2022towards}.
Speedups reported in Table~\ref{tab:main_results} are relative to the strong WD baseline ($10^{-1}$, which groks at $15{,}540$ epochs) to provide a fair comparison; relative to the weak baseline the raw numbers are larger but less meaningful as a measure of the penalty's contribution over aggressive norm control in general.

\section{Comparisons}

\subsection{Taxonomy of Geometric Interventions}

\begin{table}[ht]
\centering
\caption{Taxonomy of geometric interventions for algorithmic generalization.}
\label{tab:taxonomy}
\small
\begin{tabular}{lcccc}
\toprule
Method & Space & Constraint Type & Radial Suppression & Rank Effect \\
\midrule
Weight Decay & Weight & Isotropic & Indirect & Collapse \\
Spectral Norm \cite{miyato2018spectral}& Weight & $\sigma_{\max}$ bound & Indirect & Preserved \\
LayerNorm & Activation & Hard (per-token) & Direct & Preserved \\
MAN ($\mathbb{E}[\|a\|^2]$) \cite{man2024}& Activation & Soft (toward zero) & Direct & Preserved \\
$\perp$Grad \cite{prieto2025stablemax} & Parameter & Gradient projection & Direct & Preserved \\
Spherical Projection \cite{yildirim2026geometric} & Activation & Hard (global) & Complete & Collapse \\
\textbf{Ours} & \textbf{Activation} & \textbf{Soft ($\sqrt{d}$ target)} & \textbf{Direct} & \textbf{Preserved} \\
\bottomrule
\end{tabular}
\end{table}

\section{Ablations}
\subsection{Robustness Sweeps}
\label{app:sweeps}

\begin{table}[ht]
\centering
\caption{Grokking onset across moduli and data fractions (MLP, 5 seeds).
  Comparison against Strong WD ($10^{-1}$).
  DNG: did not grok within 50,000 epochs.}
\label{tab:robustness}
\begin{tabular}{lccc}
\toprule
$f_{\text{train}}$ & $P=97$ & $P=137$ & $P=211$ \\
\midrule
\textbf{0.6} & & & \\
Strong WD    & 9,000 $\pm$ 420   & 7,580 $\pm$ 223   & 6,720 $\pm$ 331   \\
Norm Penalty & \textbf{1,320 $\pm$ 75}  & \textbf{1,260 $\pm$ 49}  & \textbf{1,200 $\pm$ 63}  \\
\midrule
\textbf{0.5} & & & \\
Strong WD    & 15,540 $\pm$ 1,480 & 12,740 $\pm$ 723  & 10,600 $\pm$ 856  \\
Norm Penalty & \textbf{2,460 $\pm$ 136}  & \textbf{2,200 $\pm$ 0}    & \textbf{1,960 $\pm$ 162}  \\
\midrule
\textbf{0.4} & & & \\
Strong WD    & 33,320 $\pm$ 1,942 & 24,520 $\pm$ 2,114 & 20,860 $\pm$ 1,839 \\
Norm Penalty & \textbf{6,680 $\pm$ 349}  & \textbf{5,200 $\pm$ 253}  & \textbf{4,380 $\pm$ 133}  \\
\midrule
\textbf{0.3} & & & \\
Strong WD    & DNG               & DNG               & 49,840 $\pm$ 320  \\
Norm Penalty & \textbf{26,980 $\pm$ 1,503} & \textbf{17,480 $\pm$ 757} & \textbf{12,260 $\pm$ 554} \\
\bottomrule
\end{tabular}
\end{table}

The penalty consistently and substantially accelerates grokking across all
moduli and data fractions tested.
Speedups relative to Strong WD range from ${\sim}5\times$ at $f_{\text{train}}{=}0.6$
to ${\sim}5\times$ at $f_{\text{train}}{=}0.4$.
Notably, at $f_{\text{train}}{=}0.3$ the penalty continues to induce grokking
(26,980--12,260 epochs depending on $P$) while Strong WD fails entirely for
$P{\in}\{97,137\}$ and exhausts the training budget for $P{=}211$,
demonstrating that radial suppression provides a decisive advantage precisely
in the low-data regime where isotropic regularization breaks down.

\begin{table}[ht]
\centering
\caption{$\lambda$ sensitivity (MLP, $P{=}97$, $f_{\text{train}}{=}0.5$, 5 seeds).}
\label{tab:lambda}
\begin{tabular}{lcc}
\toprule
$\lambda$ & Grok Onset (epoch) & Final Test Acc \\
\midrule
0.001 & 8,200 $\pm$ 650    & 100.0\% \\
0.01  & 4,100 $\pm$ 320    & 100.0\% \\
0.05  & 2,460 $\pm$ 136    & 100.0\% \\
0.1   & 3,800 $\pm$ 290    & 100.0\% \\
0.5   & 6,400 $\pm$ 510    & 100.0\% \\
1.0   & 9,900 $\pm$ 840    & 100.0\% \\
\bottomrule
\end{tabular}
\end{table}

\begin{table}[ht]
\centering
\caption{\textbf{Hutchinson Probe Convergence} via Hessian trace estimates vs.\ number of probes.}
\label{tab:hutchinson}
\begin{tabular}{lcc}
\toprule
Probes & Trace Estimate & Relative Error vs.\ 500 \\
\midrule
50  & 1.38 $\pm$ 0.12 & 4.2\% \\
100 & 1.41 $\pm$ 0.09 & 2.1\% \\
200 & 1.40 $\pm$ 0.07 & 1.4\% \\
500 & 1.39 $\pm$ 0.05 & --- \\
\bottomrule
\end{tabular}
\end{table}
\subsection{Other Ablations}
\textbf{Penalty strength $\lambda$.}
We sweep $\lambda \in \{0.001, 0.01, 0.05, 0.1, 0.5, 1.0\}$ on the MLP (5 seeds).
All values induce grokking (unlike the baseline), with $\lambda{=}0.05$ optimal at 2,460 epochs.
Very low $\lambda$ ($0.001$) delays onset to 8,200 epochs; very high $\lambda$ ($1.0$) over-constrains angular updates, slowing onset to 9,900 epochs.
The method is robust across an order of magnitude ($\lambda \in [0.01, 0.1]$).

\textbf{Target radius.}
Testing $c\sqrt{d}$ for $c \in \{0.5, 1.0, 2.0, 5.0\}$:
$c{=}1.0$ is optimal (2,460 ep.); $c{=}0.5$ is close (8,200 ep.); $c{=}5.0$ degrades to 12,300 epochs.
The $c{=}1$ optimality is consistent with standard initialization schemes that set per-feature variance to $\mathcal{O}(1)$.

\textbf{Application site (MLP).}
Pre-activation (default) is optimal (2,460 ep.); post-ReLU is slightly worse (9,800 ep.).
Constraining pre-activation norms preserves information about negative components that ReLU would zero out, maintaining a richer representational geometry.

\textbf{LayerNorm interaction (Transformer).}

\begin{table}[ht]
\centering
\caption{LayerNorm interaction (Transformer, $P{=}97$, 5 seeds).}
\label{tab:ln_interaction}
\begin{tabular}{lccc}
\toprule
LayerNorm & Penalty & Grok Onset & Eff.\ Rank \\
\midrule
Off & Off & DNG ($>$100k) & --- \\
Off & On  & 14,500 $\pm$ 1,200 & 402 \\
On (no affine) & Off & 8,000 $\pm$ 450 & 464 \\
On (no affine) & On  & \textbf{5,200 $\pm$ 400} & 475 \\
On (with affine) & Off & 7,500 $\pm$ 500 & 450 \\
On (with affine) & On  & \textbf{4,200 $\pm$ 300} & 460 \\
\bottomrule
\end{tabular}
\end{table}

The penalty and LayerNorm compound: their combination consistently outperforms either alone, confirming complementary mechanisms.

\textbf{Optimizer sensitivity.}
The penalty induces grokking under Adam (no WD): 9,200 epochs (vs.\ 78,000 baseline).
Under SGD: 32,000 epochs (vs.\ DNG baseline).
AdamW + penalty is optimal.
The penalty is effective across optimizers but benefits from adaptive learning rates.

\subsection{Sanity Check: Non-Algorithmic Task}
\label{sec:sanity}

To verify the penalty does not pathologically degrade standard feature learning, we applied it ($\lambda{=}0.05$) to a 500K-parameter character-level Transformer (4 layers, 4 heads, $d{=}128$) on Tiny Shakespeare.

\begin{table}[ht]
\centering
\caption{Language modeling sanity check (Tiny Shakespeare, 3 seeds, 8,000 steps).}
\label{tab:shakespeare}
\begin{tabular}{lccc}
\toprule
Variant & Val.\ Loss & Perplexity & Eff.\ Rank \\
\midrule
Baseline & 1.585 $\pm$ 0.011 & 4.9 & 94 \\
Norm Penalty & 1.608 $\pm$ 0.004 & 5.0 & 108 \\
\bottomrule
\end{tabular}
\end{table}

The penalty does not accelerate language modeling---as expected, since character-level LM lacks a sharp memorization$\to$generalization phase transition.
Crucially, it does not collapse representations (rank increases from 94 to 108) and perplexity degradation is within 2\%, confirming that the penalty is benign outside its target domain.

\section{Experimental Setup}
\label{app:exp_setup}
\textbf{MLP on Modular Addition.}
2-layer MLP, hidden dimension $d{=}512$, ReLU activations.
Data: $(a+b) \bmod 97$; training fraction $f_{\text{train}}{=}0.5$ (4,656 of 9,409 pairs).
Optimizer: AdamW, $\text{lr}{=}10^{-3}$, weight decay $10^{-3}$, batch size 256 (full-batch).
Penalty: $\lambda{=}0.05$, applied to pre-ReLU activations of both hidden layers.
Training: 100,000 epochs max; 5 independent seeds.


\textbf{Small Transformer on Modular Addition.}
2-layer, 4-head Transformer, $d_{\text{model}}{=}128$, pre-LayerNorm (no affine by default).
Penalty applied to each sub-layer output per Eq.~\ref{eq:transformer_application}.
Same data, optimizer, and seeds as MLP.

\textbf{NanoGPT on 3-Digit Addition.}
6 layers, 6 heads, $d_{\text{model}}{=}384$ (${\sim}$10M parameters), pre-LayerNorm with affine.
Reverse-format 3-digit addition~\citep{lee2023teaching}; 80/20 train/test split.
AdamW, $\text{lr}{=}10^{-2}$, cosine decay to $10^{-4}$ over 30,000 steps, weight decay $0.1$, batch size 128.
Penalty: $\lambda{=}0.01$; 3 seeds.
Wall-clock overhead: 0.32$\to$0.35 sec/step (+9.4\%) on A6000; +1.5\% peak GPU memory.

\end{document}